\documentclass[journal]{IEEEtran}
\usepackage{graphicx}
\usepackage{subfigure}
\usepackage[square, comma, sort&compress, numbers]{natbib}
\usepackage{times}
\usepackage{latexsym}

\usepackage{url}
\usepackage{color}
\usepackage{microtype}
\usepackage{booktabs} 
\usepackage{epstopdf}
\usepackage{epsfig}

\usepackage{amsfonts}
\usepackage{amsmath}
\usepackage{amssymb}
\usepackage{bm}
\usepackage{dsfont}

\usepackage{algorithmic}
\usepackage{algorithm,times}

\usepackage{threeparttable}
\usepackage{multirow}
\usepackage{tabularx}
\usepackage{array}
\newcommand{\PreserveBackslash}[1]{\let\temp=\\#1\let\\=\temp}
\newcolumntype{C}[1]{>{\PreserveBackslash\centering}p{#1}}
\newcolumntype{R}[1]{>{\PreserveBackslash\raggedleft}p{#1}}
\newcolumntype{L}[1]{>{\PreserveBackslash\raggedright}p{#1}}

\begin{document}
%
\title{Automated Essay Scoring based on Two-Stage Learning}

\author{
	\IEEEauthorblockN{Jiawei Liu$^{\dag}$, Yang Xu$^{\ddag}$, and Yaguang Zhu$^{\dag}$\\
	\IEEEauthorblockA{$^{\dag}$Fopure (Hangzhou) Technology Co., Ltd., Hangzhou, China}\\
	\IEEEauthorblockA{$^{\ddag}$School of Computer Science and Technology, University of Science and Technology of China, Hefei, China}\\
    Email: liujiawei@fopure.com, smallant@mail.ustc.edu.cn, zhuyaguang@fopure.com
    }
}

\maketitle

\begin{abstract}
Current state-of-the-art feature-engineered and end-to-end Automated Essay Score (AES) methods are proven to be unable to detect adversarial samples, e.g. the essays composed of permuted sentences and the prompt-irrelevant essays. 
Focusing on the problem, we develop a Two-Stage Learning Framework (TSLF) which integrates the advantages of both feature-engineered and end-to-end AES methods. 
In experiments, we compare TSLF against a number of strong baselines, and the results demonstrate the effectiveness and robustness of our models. 
TSLF surpasses all the baselines on five-eighths of prompts and achieves new state-of-the-art average performance when without negative samples. 
After adding some adversarial eassys to the original datasets, TSLF outperforms the features-engineered and end-to-end baselines to a great extent, and shows great robustness.
\end{abstract}


\IEEEpeerreviewmaketitle

\section{Introduction}
\label{intro}
Automated Essay Scoring (AES), which extracts various features from essays and then scores them on a numeric range, can improve the efficiency of writing assessment and reduce human efforts to a great extent. 
In general, AES models can be divided into two main streams. 
The models of the first stream are feature-engineered models, which are driven by handcrafted features, such as the number of words and grammar errors \cite{Yannakoudakis2011A,Chen2014Automated}.
The advantage is that the handcrafted features are explainable and flexible, and could be modified and adapted to different scoring criterion. 
However, some deep semantic features extracted by understanding the essays, which are especially essential for prompt-dependent writing tasks, are hard to be captured by feature-engineered models.

The other stream is the end-to-end model, which is driven by the rapid development of deep learning techniques \cite{Taghipour2016A,Alikaniotis2016Automatic,farag2018neural,jin2018tdnn}. 
Specifically, based on word embedding \cite{mikolov2013distributed,pennington2014glove}, essays are represented into low-dimensional vectors, and followed by a dense layer to transform these deep-encoded vectors (involving deep semantic meanings) to corresponding ratings. 
Although end-to-end models are good at extracting deep semantic features, they can hardly integrate the handcrafted features like spelling errors and grammar errors, which are proven to be vital for the effectiveness of AES models. 
In this paper, we argue that both handcrafted features and deep-encoded features are necessary and should be exploited to enhance AES models.

\begin{figure*}[ht]
	\begin{center}
		\centerline{\includegraphics[width=0.9\textwidth]{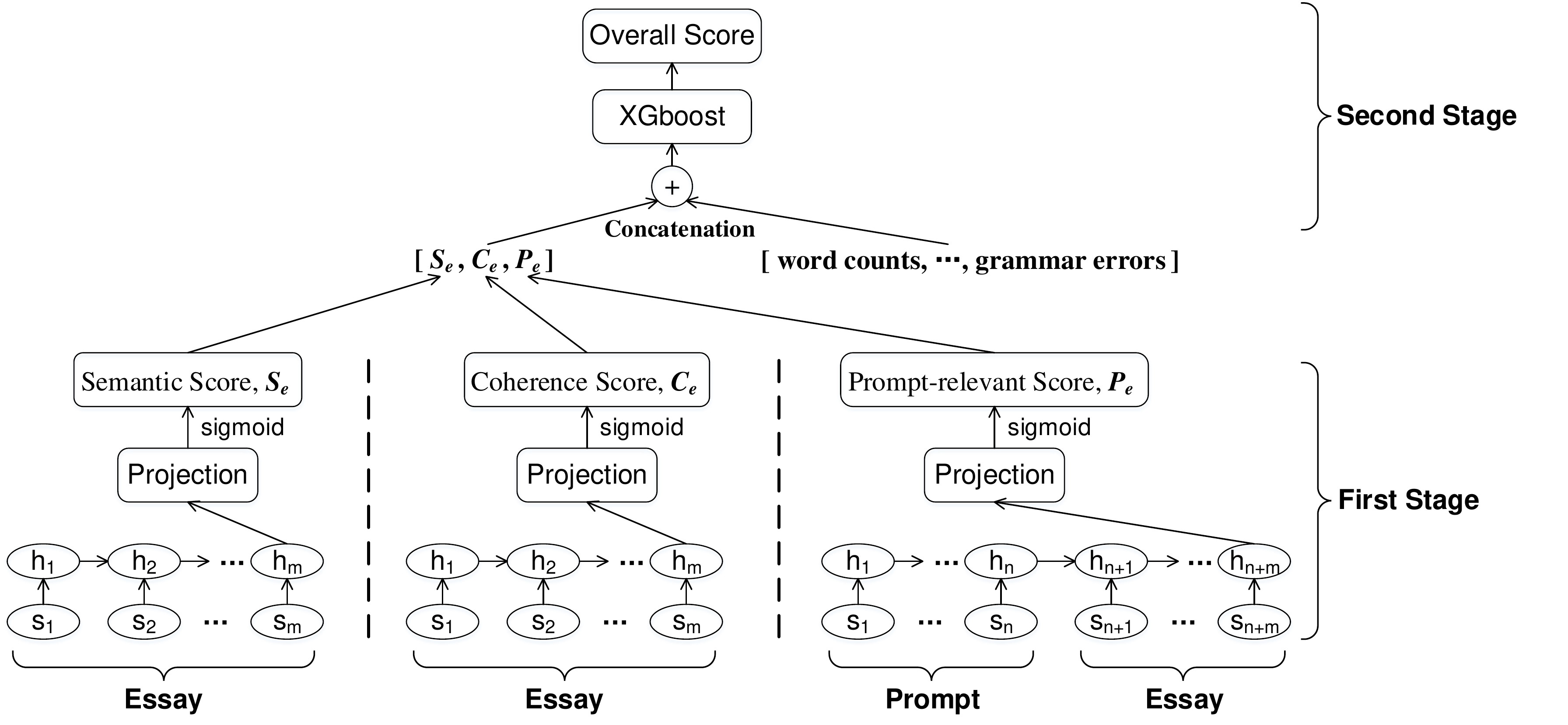}}
		\caption{Two-Stage Learning Framework for AES. In the first stage, based on deep neural networks, we calculate semantic score, coherence score and prompt-relevant score named as $S_e$, $C_e$ and $P_e$ respectively. $C_e$ and $P_e$ are proposed to detect adversarial samples. In the second stage, we concatenate these three scores with some handcrafted features and feed the result to the boosting tree model for further training. }
		\label{TSLF}
	\end{center}
\end{figure*}

It is reported that some well-designed adversarial inputs can be exploited to cheat AES models so that the writers who are familiar with the systems' working can maximize their scores \cite{Powers2002Stumping}. 
Generally, there are two categories of adversarial inputs. 
One is composed of well-written permuted paragraphs, which have been successfully detected by \cite{farag2018neural} based on a coherence model \cite{li2014model}. 
The other consists of prompt-irrelevant essays, which remain to be dealt with. 
Focusing on the problems and arguments mentioned above, in this paper, we develop a Two-Stage Learning Framework (TSLF), which makes full use of the advantages of  feature-engineered and end-to-end methods. 
In the first stage, we calculate three scores including semantic score, coherence score and prompt-relevant score based on Long Short-Term Memory (LSTM) neural network. Semantic score is prompt-independent, and utilized to evaluate essays from deep semantic level.
Coherence score is exploited to detect the essays composed of permuted paragraphs.
The connections between prompts and essays are evaluated based on prompt-relevant scores, which are defined to detect the prompt-irrelevant samples. 
In the second stage, we concatenate these three scores with some handcrafted features, and the results are fed the eXtreme Gradient Boosting model (XGboost) \cite{DBLP:journals/corr/ChenG16} for further training. The details of TSLF are illustrated in Figure \ref{TSLF}.
In experiments, TSLF together with a number of strong baselines are evaluated on the public Automated Student Assessment Prize dataset (ASAP) \cite{Taghipour2016A}, which consists of $8$ prompts. 
Our contributions in this paper are summarized as follows.
\begin{itemize}
	\item  The results on the original ASAP dataset demonstrate the effectiveness of integrating both feature-engineered models' advantages and end-to-end models' advantages. TSLF outperforms the baselines on five-eighths of prompts and achieves new state-of-the-art performance on average.
	\item  After adding some adversarial samples to the original ASAP dataset, TSLF surpasses all baselines to a great degree, and show great robustness. The results demonstrate the validity of our coherence model and prompt-relevant model to detect the negative samples.
	\item With respect to the handcrafted features, the current AES models only concern about the spell errors. However, other grammar errors such as article errors and preposition errors are also very important for a valid AES system. To the best of our knowledge, we are the first to introduce the Grammar Error Correction (GEC) system into AES models. 
\end{itemize}

\section{Two-Stage Learning Framework (TSLF)}
\label{model} 
In Section \ref{intro}, we argue that deep-encoded features and handcrafted features are both necessary for a valid AES system. 
In this section, we are going to introduce our Two-Stage Learning Framework (TSLF), which combines the advantages of feature-engineered models and end-to-end models.
As shown in Figure \ref{TSLF}, during the first stage, we calculate semantic score $S_e$, coherence score $C_e$ and prompt-relevant score $P_e$, where $C_e$ is utilized to detect the adversarial samples composed of well-written permuted paragraphs and $P_e$ is designed for prompt-irrelevant samples. 
In the second stage, these three scores together with some handcrafted features are concatenated and fed to a boosting tree model for further training.

\subsection{Sentence Embedding}
It is proven that the context-dependent embedding method named Bidirectional Encoder Representations from Transformers (BERT) achieved new state-of-the-art results on some downstream tasks like question answering and text classification \cite{devlin2018bert}. 
Due to these exciting achievements, in this paper, sentence embeddings are derived by the pre-trained BERT model\footnote{https://github.com/google-research/bert. In this paper, we utilize the uncased model with 12-layer, 768-hidden, 12-heads and 110M parameters.}. For sentence $s =\{\rm t_0,t_1, \cdots, t_n, t_{n+1}\}$, where $\rm t_i$ $(0 \leq i \leq n+1)$ indicates the $i^{th}$ word in sentence, $\rm t_0$ is a special tag $\rm CLS$ used for classification tasks and $\rm t_{n+1}$ is another special tag $\rm SEP$ utilized to split the sentences. Every word in the sentence including $\rm CLS$ and $\rm SEP$ will be encoded into a low-dimensional embedding $w_i   (w_i \in \mathbb{R}^d)$ based on BERT. 
In this paper, the average of the hidden states of the penultimate transformer layer along the time axis is exploited to represent the sentence.
Concretely, the representation of sentence $s$ is expressed as the following equation.
\begin{equation}
s_{snt} = \frac{1}{n+2}\sum\limits_{i=0}^{n+1} w_i^{-2}
\end{equation}
where $s_{snt}$ means a sentence's embedding, superscript $-2$ of $w_i$ indicates word representations are from the penultimate transformer layer.  
In this paper, we do not make use of the last layer's representations because the last layer is too closed to the target functions of pre-training tasks including masked language model task and next sentence prediction task \cite{devlin2018bert}, and therefore the representations may be biased to those targets.

\subsection{First Stage} 
\textbf{Semantic Score}
In the first stage, we utilize LSTM to map essays into low-dimensional embeddings, which are then fed to a dense output layer for scoring essays. Concretely, for an essay $e=\{s_1, s_2, \cdots, s_m\}$, where $s_t$ indicates the $t^{th}$ $ (1 \leq t \leq m, s_t \in \mathbb{R}^d)$ sentence embedding in the essay and $d$ means the length of sentence embedding. The encoding process of LSTM is described as follows:
\begin{equation}
\begin{array}{ll}
i_t = &\sigma(W_i \cdot s_t + U_i \cdot h_{t-1} + b_i) \\
f_t = &\sigma(W_f \cdot s_t + U_f \cdot h_{t-1} + b_f) \\
\tilde{c}_t = &\sigma(W_c \cdot s_t + U_c \cdot h_{t-1} + b_c) \\
c_t = &i_t \circ \tilde{c}_t + f_t \circ c_{t-1} \\
o_t = &\sigma(W_o\cdot s_t + U_o \cdot h_{t-1} + b_o) \\
h_t = &o_t \circ tanh(c_t)
\end{array}
\end{equation}
$h_t$ means the hidden state of sentence $s_t$. $W_i$, $W_f$, $W_c$, $W_o$, $U_i$, $U_f$, $U_c$, $U_o$ are the weight matrices for the input gate, forget gate, candidate state, and output gate respectively. $b_i$, $b_f$, $b_c$, $b_o$ stand for the bias vectors. $\sigma$ denotes the sigmoid function and $ \circ $ means element-wise multiplication. Hence, for the essay $e$, we will get the hidden state set $H = \{h_1, h_2, \cdots, h_m\}$. In this paper, we utilize the last hidden state rather than the average hidden state \cite{Taghipour2016A, farag2018neural} to define the final essay's representation. In Section \ref{experiments}, we demonstrate that the average hidden state does not perform as well as the last hidden state. 
Essay representation $h_{m}$ is then fed into a dense layer to transform the low-dimensional vector into a scalar value.
However, different writing tasks may have different score ranges. For instance, in the writing task of the Test of English as a Foreign Language (TOEFL), score ranges from $0$ to $30$. Differently, the score range of the International English Language Testing System (IELTS) is $(0,9)$. 
Since a fixed score range is necessary for different writing tasks, hence, we project the output of dense layer to a scaled value in the range $(0, 1)$ by utilizing the sigmoid function. Concretely, the equation is defined as follows:
\begin{equation}
\label{semantic score}
S_e = sigmoid(w_s \cdot h_{m} + b_s)
\end{equation}
$S_e$ indicates the semantic score of essay $e$. $w_s$ means the weighted matrix of the dense layer and $b_s$ stands for the bias. 
The objective we choose is the Mean Squared Error (MSE) loss function.  Given the training set with $N$ samples $E=\{e_1, e_2, \cdots, e_N\}$, the objective is described as Equation \ref{objective semantic}. 
\begin{equation}
\label{objective semantic}
obj(S_E, \tilde{S}_E) = \frac{1}{N}\sum\limits_{i=1}^N(S_i -  \tilde{S}_i)^2
\end{equation}
where $S_E$ means the predict score set of training samples, $\tilde{S}_E$ indicates the original hand marked score set.

\textbf{Coherence Score}
For the first kind of adversarial samples, the essays containing permuted well-written paragraphs, we also utilize the coherence model to detect. Obviously, coherence scores for well-organized essays must be higher than the permuted essays. Differently, our coherence model is LSTM-based rather than utilizes the clique strategy mentioned in \cite{li2014model,farag2018neural}. Details of the coherence model are illustrated in Figure \ref{TSLF}. Obviously, the structure of our coherence model is the same as the model utilized to get the semantic score. Similarly, for an essay $e=\{s_1, s_2, \cdots, s_m\}$, the final coherence score is defined as Equation \ref{coherence score}. 
\begin{equation}
\label{coherence score}
C_e = sigmoid(w_c \cdot h_{m} + b_c)
\end{equation}
where $C_e$ denote the coherence score, $w_c$ indicates the weighted matrix, $b_c$ is the bias and $h_{m}$ is the final hidden state of LSTM. The objective of our coherence model is also built by exploiting the MSE, which is shown as follows.
\begin{equation}
\label{objective coherence}
obj(C_E, \tilde{C}_E) = \frac{1}{N}\sum\limits_{i=1}^N(C_i -  \tilde{C}_i)^2
\end{equation}
$C_E$ stands for the predict coherence score set of training samples, $\tilde{C}_E$ denotes the gold coherence score set and $N$ is the number of training samples.
During the training process, we assume that the gold coherence scores of the original essays are equal to the corresponding hand marked scores, and the assumption is also adopted by \cite{farag2018neural}. For these permuted essays, the gold coherence scores are set as $0$.

\textbf{Prompt-relevant Score} 
The prompt-relevant score is calculated based on LSTM again, and the details are illustrated in Figure \ref{TSLF}. For a specific prompt composed of $n$ sentences $p=\{s_1, s_2, \cdots, s_n\}$ and an essay $e=\{s_1, s_2, \cdots, s_m\}$ with $m$ sentences, we first combine $p$ and $e$, and the combination is defined as $\tilde{e}=\{s_1, s_2, \cdots, s_{m+n}\}$. $s_i (1<i<m+n)$ denotes the sentence embedding.
By exploiting LSTM, sentence set $\tilde{e}$ is encoded to a hidden state set $H=\{h_1, h_2, \cdots, h_{m+n}\}$. Then, the last hidden state $h_{m+n}$ is fed to a non-linear followed by sigmoid activation to get the prompt-relevant score. The Equation is defined as follows. 
\begin{equation}
\label{prompt-relevant score}
P_e = sigmoid(w_p \cdot h_{m+n} + b_p)
\end{equation}
where $P_e$ is prompt-relevant score, $w_p$ and $b_p$ indicates the weighted matrix and bias respectively. The objective is again base on MSE and shown in Equation \ref{objective prompt-relevant}.
\begin{equation}
\label{objective prompt-relevant}
obj(P_E, \tilde{P}_E) = \frac{1}{N}\sum\limits_{i=1}^N(P_i - \tilde{P}_i)^2
\end{equation}
$P_E$ stands for the predict prompt-relevant score set of training samples, $\tilde{P}_E$ denotes the gold prompt-relevant score set and $N$ is the number of training samples. In the training set, the gold scores of prompt-relevant essays are assumed as the hand marked scores. For prompt-irrelevant essays, the gold prompt-relevant scores are defined as $0$.

\subsection{Second Stage}
We argue that handcrafted features are also necessary for the improvement of AES systems. In this stage, $S_e$, $C_e$ and $P_e$ together with some handcrafted features are applied for further training. Score $S_e$ contains semantic-related information. $C_e$ and $P_e$ are exploited to antagonize the adversarial samples.

\begin{table}[h]
	\caption{Handcrafted Features used in Our Two-Stage AES model}
	\label{handcrafted}
	\begin{center}
		\begin{small}
			\begin{sc}
				\begin{tabularx}{\columnwidth}{cX} 
					\toprule
					Index & \multicolumn{1}{c}{Feature Description}\\
					\midrule
					1    & Number of grammar Errors \\
					2    & Essay length in words and Characters \\
					3    & Mean and variance of the word length \\
					& in Characters \\
					4    & Mean and variance of sentence length \\
					& in words \\
					5    & Number of Clause in an essay \\
					6    & Vocabulary size in essay \\
					\bottomrule
				\end{tabularx}
			\end{sc}
		\end{small}
	\end{center}
\end{table}

\textbf{Handcrafted Features} 
Some of the handcrafted features utilized in this paper are from \cite{jin2018tdnn} and shown in Table \ref{handcrafted}. Intuitively, essays with more grammar errors tend to be assigned with lower scores. In addition, we observe that essays with the lowest scores usually contains much fewer words and sentences than other essays. Hence, length-based features are taken into consideration. Complexities of words and sentences are evaluated in terms of corresponding mean and variance of word length and sentence length. Besides, clause number is another aspect, which indicates the grammar ability of writers. Vocabulary size i.e., the number of unique words in an essay, is utilized to evaluate writers' vocabulary level.

\textbf{Grammar Error Correction (GEC)}
As what we investigate, traditional feature-engineered AES methods only concern spell errors. However, other grammar errors such as preposition errors and article errors are also necessary for evaluating essays. Hence, a valid GEC system is necessary. Actually, in the field of GEC, many distinguished achievements have been made.  \cite{junczysdowmunt-grundkiewicz} trained a statistic machine translation model for this task and got the state-of-the-art performance by exploiting Moses\footnote{http://www.statmt.org/moses/}. \cite{ge2018fluency} proposed three boost learning methods based on traditional sequence-to-sequence framework, and achieved a new state-of-the-art result. Since GEC is not the dominated part in this paper, we adapt the pre-trained GEC model\footnote{https://github.com/grammatical/baselines-emnlp2016} to conduct grammar error correction. The open-source software Jamspell\footnote{https://github.com/bakwc/JamSpell} is chosen to do spell check.

\textbf{XGboost Learning}  XGboost\footnote{https://github.com/dmlc/xgboost} is a well-known open-source software for its efficiency, flexibility and portability \cite{DBLP:journals/corr/ChenG16}. It provides a boosting model termed as Gradient Boosting Decision Tree (GBDT), and consists of 
two basic models including the tree model and linear model. Given handcrafted features set $H$, the overall score $O_e$ of an essay is defined as follows.
\begin{equation}
\label{xgboost}
O_e = {\rm XGboost}([H;\ S_e;\ C_e;\ P_e])
\end{equation}
$[\ ]$ means concatenating operation. 

\subsection{Training}
\label{training}
In ASAP dataset, score ranges are different with each other. For consistency, the gold scores in the training set are first normalized to the range $(0, 1)$. During the testing process, predict scores will be rescaled to the original ranges.  
For training the objectives in the first stage, we utilize Adam to update the trainable variables. Hyperparameters $\beta_1$, $\beta_2$ and $\epsilon$ are equal to $0.9$, $0.999$ and $1e-6$ respectively. Initial learning rate is set as $0.00001$. These hyperparameters are also adopted in the source code of BERT \cite{devlin2018bert}.
All LSTM neural networks used in this paper are single-layer with hidden size of $1024$. Actually, we also test the performance of multi-layer and bi-direction LSTM model. But the results are not as good as we expected. To avoid overfitting, dropout is applied in the training process and the proportion is set as $0.5$. 

During the second stage learning, we apply the tree-based boosting model and 
the learning rate is set as $0.001$. The max depth of tree is equal to $6$. Logistic function is chosen to be the objective. Other parameters are set as default.  Early stopping epochs are set as $100$. Concretely, if the loss doesn't change anymore in $100$ consecutive training rounds, we will stop training.

At each training step of the coherence model, we permute each origin essay in the training batch to generate the adversarial samples. Specifically, the size of training batch will be $32$ after we add the adversarial samples to the origin training batch who size is $16$. 
To generate prompt-irrelevant samples, for every prompt with $m$ essays in the training set, we randomly select $m$ negative samples from other prompt so that the imbalance problem can be avoid. The gold scores of these adversarial samples are set as $0$. 

\section{Experiments}
\label{experiments}
\subsection{Experimental Setup}
\textbf{Dataset and Corpus} The most widely used dataset for AES is the Automated Student Assessment Prize (ASAP)\footnote{https://www.kaggle.com/c/asap-aes}, and it has been utilized to evaluate the performance of AES systems in \cite{Alikaniotis2016Automatic, Taghipour2016A, jin2018tdnn,cozma2018automated}. Generally, ASAP is composed of $8$ prompts and $12976$ essays, which are written by students from Grade 7 to Grade 10. More details are summarized in Table \ref{asap dataset}. Similar to \cite{Taghipour2016A, farag2018neural}, we apply 5-fold cross validation on ASAP. 

\begin{table}[h]
	\caption{Details about ASAP dataset}
	\label{asap dataset}
	\begin{center}
		\begin{small}
			\begin{sc}
				\begin{tabularx}{\columnwidth}{cX<{\centering}X<{\centering}} 
					\toprule
					Prompt ID & Essays & Score Range \\
					\midrule
					1    & 1783  & 2-12 \\
					2    & 1800  & 1-6  \\
					3    & 1726  & 0-3  \\
					4    & 1772  & 0-3  \\
					5    & 1805  & 0-4  \\
					6    & 1800  & 0-4  \\
					7    & 1569  & 0-30 \\
					8    & 723   & 0-60 \\
					\bottomrule
				\end{tabularx}
			\end{sc}
		\end{small}
	\end{center}
\end{table}

\textbf{Evaluation Metric}
For consistency, the predicted scores together with the gold scores in the test set are uniformly re-scaled into $[0, 10]$. Similar to \cite{Alikaniotis2016Automatic, Taghipour2016A, jin2018tdnn, farag2018neural}, we test the performance of TSLF based on the Quadratic Weighted Kappa (QWK), which is the official evaluation metric\footnote{https://github.com/benhamner/asap-aes} and exploited to test the agreement between the predicted scores and the gold scores. 

\textbf{Strong Baselines}
The baselines are utilized to compare with TSLF and listed in Table \ref{asap result}. 
EASE (Enhanced AI Scoring Engine) is a feature-engineered AES engine (introduced in Section \ref{related work}), and it got the third place in the public competition of ASAP by using parts-of-speech (POS), n-gram and length-based features. In EASE, the extracted handcrafted features are fed into a regression model for training.
In this paper, we choose Support Vector Regression (SVR) and Bayesian Linear Ridge Regression (BLRR) \cite{phandi} for comparison.
With respect to end-to-end AES baselines, we choose three models proposed in \cite{Taghipour2016A} including CNN, LSTM and CNN+LSTM, which achieved the state-of-the-performance on ASAP. 
In CNN, all word embeddings are fed to a convolution layer to extract essays' representations. In LSTM, essays' representations are extracted by a recurrent layer rather than convolution layer. In CNN+LSTM, the outputs of convolution layer are then fed into recurrent layer for further extraction. Essays' representations are then transformed to scalars based on sigmoid activation.

\begin{table*}[ht]
	\caption{Results On ASAP Dataset without adverserial samples}
	\label{asap result}
	\begin{center}
		\begin{small}
			\begin{sc}
				\begin{tabularx}{\textwidth}{lX<{\centering}<{\centering}X<{\centering}X<{\centering}X<{\centering}X<{\centering}X<{\centering}X<{\centering}X<{\centering}X<{\centering}} 
					\toprule
					Models  & prompt1 & prompt2 & prompt3 & prompt4 & prompt5 & prompt6 & prompt7 & prompt8 & Average \\
					\midrule
					EASE(SVR)  & 0.781  & 0.621  & 0.630  & 0.749  & 0.782  & 0.771  & 0.727  & 0.534  & 0.699  \\
					EASE(BLRR) & 0.761  & 0.606  & 0.621  & 0.742  & 0.784  & 0.775  & 0.730  & 0.617  & 0.705  \\
					\hline
					CNN        & 0.804  & 0.656  & 0.637  & 0.762  & 0.752  & 0.765  & 0.750  & 0.680  & 0.726  \\
					LSTM       & 0.808  & 0.697  & 0.689  & 0.805  & 0.818  & \textbf{0.827}  & \textbf{0.811}  & 0.598  & 0.756 \\
					CNN+LSTM   & 0.821  & 0.688  & 0.694  & \textbf{0.805}  & 0.807  & 0.819  & 0.808  & 0.644  & 0.761 \\
					\hline
					TSLF-1     & 0.757  & 0.698  & 0.725  & 0.796  & 0.810  & 0.783  & 0.727  & 0.544  & 0.730  \\
					TSLF-2     & 0.808  & 0.718  & 0.693  & 0.698  & 0.771  & 0.720  & 0.722  & 0.616  & 0.718  \\
					TSLF-ALL   & \textbf{0.852}  & \textbf{0.736}  & \textbf{0.731}  & 0.801  & \textbf{0.823}  & 0.792  & 0.762  & \textbf{0.684}  & \textbf{0.773}  \\
					\bottomrule
				\end{tabularx}
			\end{sc}
		\end{small}
	\end{center}
\end{table*}

\subsection{ASAP Test without Adversarial Essays}
In this subsection, we test the performance of the baselines and TSLF on the orginal ASAP dataset. 
As mentioned above, ASAP is composed of 8 prompts. For better comparison, we report the QWK scores of baselines and TSLF on each prompt, and average the scores to evaluate the overall performance. The results are shown in Table \ref{asap result}. 
The results of EASE are from \cite{phandi}. EASE(SVR) means the extracted features of EASE are fed to SVR for regression, and EASE(BLRR) stands for that the training process is conducted based on BLRR.
The results of CNN, LSTM and CNN+LSTM are from \cite{Taghipour2016A}, which achieved the state-of-the-art performance on ASAP. 
TSLF-1, TSLF-2, TSLF-ALL represents our models. TSLF-1 is similar to other end-to-end models and utilizes the semantic score as the overall essay's score.
TSLF-2 is actually a feature-engineer model, which only utilize the handcrafted features proposed in Table \ref{handcrafted} to predict essay's score. TSLF-ALL denotes the whole pipeline of TSLF. Since there are no adversarial essay in ASAP, during the second stage learning, we remove the coherence score and prompt-relevant score from the concatenated features in Equation \ref{xgboost}. 

\textbf{Analysis}
From Table \ref{asap dataset}, we observe that TSLF-ALL outperforms all baselines in five-eighths of prompts and leads the average performance, which is consistent with our expectation. 
After analyzing the ASAP dataset, we note that the samples whose scores are much lower than the average usually are very short essays. Besides, a lot of essays have grammar errors and spell errors. From our perspective, these length-based features and errors are difficult for the end-to-end models to make full use of. However, according to the second stage learning, TSLF-ALL can take advantage of not only the deep-encoded features generated by end-to-end models but also the handcrafted features, which accounts for the impressive achievements of TSLF-ALL. 
In addition, with respect to the feature-engineered model in Table \ref{asap dataset}, we find that TSTF-2 gets better performance on average but uses fewer features than EASE. We speculate that the improvements of TSTF-2 may credit to the introduction of GEC system and the advanced ability of the tree boosting regression model. 
\begin{figure}[ht]
	\begin{center}
		\centerline{\includegraphics[width=\linewidth]{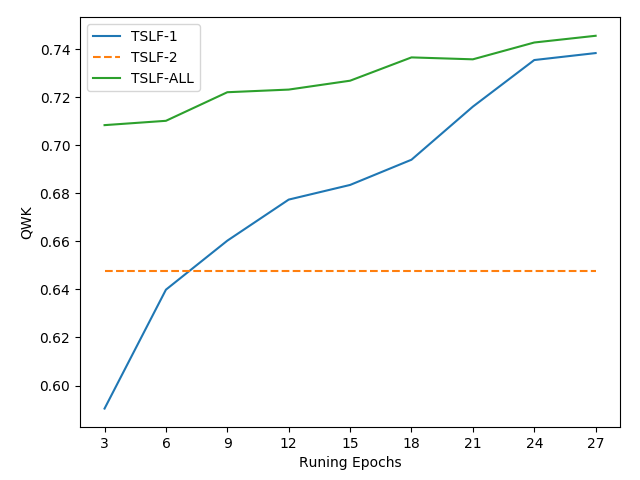}}
		\caption{The relationship between TSLF-1, TSLF-2 and TSLF-ALL}
		\label{connections}
	\end{center}
\end{figure}

Furthermore, it shows that TSLF-ALL always performs much better than TSLF-1 and TSLF-2.
To better understand the connections between TSLF-1, TSLF-2 and TSLF-All, we shuffle the whole ASAP dataset first, and $80\%$ essays is used for training and the left is applied for testing. During each 3 training epochs of TSLF-1 and TSLF-ALL, we calculate the QWK scores on the test data and the results are illustrated in Figure \ref{connections}, where the blue solid line indicates TSLF-1, the orange dash line denotes TSLF-2 and the green solid line represents TSLF-ALL. We note that the results in Figure \ref{connections} are consistent with our observations in Table \ref{asap result}. Moreover, the performance of TSLF-ALL shows an ascending trend and always keeps the pace with TSLF-1. In addition, TSLF-ALL always performs much better than the other two models especially at the beginning of training process,  which demonstrates that the performance of AES can benefit from integrating both deep-decoded features and handcrafted features. 
\begin{figure}[ht]
	\begin{center}
		\centerline{\includegraphics[width=\linewidth]{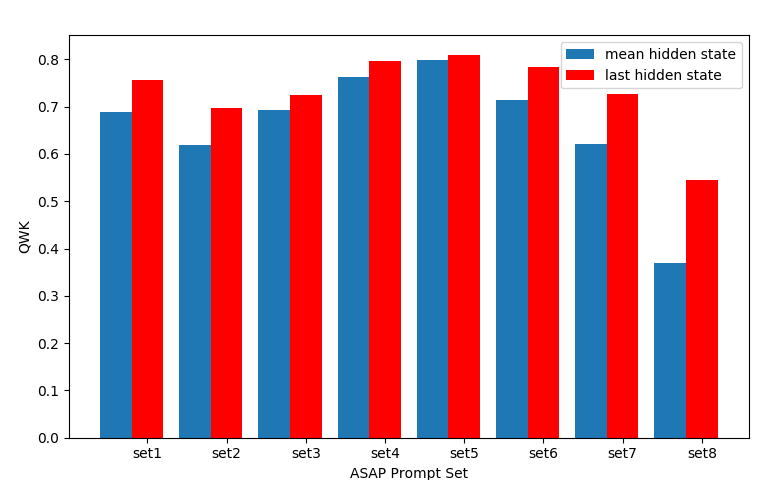}}
		\caption{The performance of difference essay's representations.}
		\label{meantime-vs-lasttime}
	\end{center}
\end{figure}

In \cite{Taghipour2016A, farag2018neural}, the essay representation is calculated by a mean-over-time operation on the LSTM's hidden states. Actually, essay's representation can also be obtained by using the last hidden state. Hence, we compare the performance of the two kinds of essay's representation. The results are illustrated in Figure \ref{meantime-vs-lasttime}. It is explicit that the last hidden state is superior to the average hidden state.  
In this paper, the semantic score, coherence score and prompt-relevant score are all calculated by the last hidden state of LSTM due to its efficiency.

\subsection{ASAP Test with Adversarial Essays}
In this subsection, we test the ability of AES system to detect the adversarial essays. As mentioned above, we utilize a coherence score to detect the well-written permuted paragraphs, and apply prompt-relevant score to detect the prompt-irrelevant essays.

\begin{table}[h]
	\caption{Results On ASAP Dataset with adverserial samples}
	\label{asap dataset with adver}
	\begin{center}
		\begin{small}
			\begin{sc}
				\begin{tabularx}{\columnwidth}{lccc} 
					\toprule
					Methods    & Adver1 & Adver2 & Adver1+Adver2 \\
					\midrule
					TSLF-1     & 0.119  & 0.169  & 0.160    \\
					TSLF-2     & 0.128  & 0.178  & 0.060    \\
					TSLF-ALL   & \textbf{0.651}  & \textbf{0.883}  & \textbf{0.709} \\
					\bottomrule
				\end{tabularx}
			\end{sc}
		\end{small}
	\end{center}
\end{table}

\textbf{Adversarial Essays}
Generally, there are three possible ways to add adversarial samples to original ASAP dataset: 
(1) We only add the permuted essays to ASAP.
(2) We only add the prompt-irrelevant essays to ASAP. 
(3) We add both permuted essays and prompt-irrelevant essays to ASAP. 
In condition 1, for every essay in ASAP dataset, we permute sentences to generate the corresponding adversarial essay. 
In condition 2, we randomly select the same number of essays from other prompt set to generate prompt-irrelevant samples.
In condition 3, we create the same number of permuted and prompt-irrelevant essays respectively and make sure that the whole number of negative samples are equal to the number of gold essays in a prompt.
We report the average performance on ASAP by using QWK scores, and the results are listed in Table \ref{asap dataset with adver}.

\begin{figure}[ht]
	\begin{center}
		\centerline{\includegraphics[width=\linewidth]{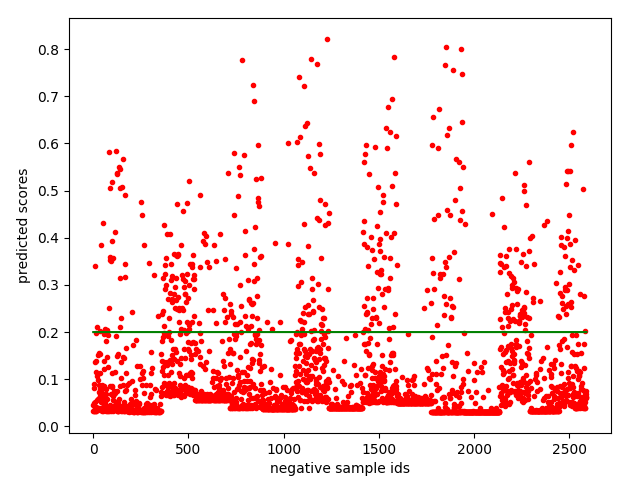}}
		\caption{The performance of difference essay's representations.}
		\label{negative samples}
	\end{center}
\end{figure}

\textbf{Analysis}
From Table \ref{asap dataset with adver}, we observe that the end-to-end model TSLF-1 and feature-engineered model TSLF-2 are weak when adversarial essays are added to the test set, which is consistent with our expectations. 
However, TSLF-ALL is much more robust because of the addition of coherence score and prompt-relevant score.
Besides, we find that the permuted essays are much more difficult to be detect than the prompt-irrelevant essays. Actually, in experiments, the prompt-relevant model only needs $3$ epochs to converge while the coherence model converges after $120$ epochs. Obviously, there are massive permuted samples for a specific essay, which are added to the training batch in every epoch. From our perspective, the much more training epochs are caused by the large number of permuted essays. To better understand the robustness of our model, we illustrate the predicted overall scores of negative samples in Figure \ref{negative samples}.
The red points denote the predicted scores and more than $82\%$ of the points are below $0.2$ (green line). Obviously, most of the negative samples including the permuted essays and prompt-irrelevant essays are assigned with extremely low marks. Therefore, we argue that the coherence model and prompt-relevant model are valid and effective for improving the robustness of AES systems.

\section{Related Work}
\label{related work}
\textbf{Automated Essay Scoring} As we mentioned above, AES models are divided into feature-engineered models and end-to-end models. With regard to the first category, \cite{Yannakoudakis2011A, Chen2014Automated, jin2018tdnn} trained a Rank Support Vector Machine (RankSVM) to predict essay scores by utilizing pre-defined handcrafted features. Besides, the open-source AES system\footnote{https://github.com/edx/ease} named Enhanced AI Scoring Engine (EASE) applied several regression methods to score essays on the basis of  handcrafted features.
As for the end-to-end AES methods, \cite{Alikaniotis2016Automatic} encoded an essay into a low-dimensional embedding by exploiting a deep LSTM network. At the output layer of their model, the essay embeddings was fed to predict the essay score through a linear unit. It was reported that their approach outperformed the feature-engineered models on the ASAP dataset. \cite{Taghipour2016A} first utilized Convolutional Neural Networks (CNN) to extract context features from word level. The outputs of the CNN layers were then fed to a followed LSTM layer to get the final essay embedding. Again, a linear unit was leveraged to map the essay embedding to a specific score. Similar to the CNN+LSTM model proposed in \cite{Taghipour2016A}, \cite{farag2018neural} removed the CNN layer and essay features were directly extracted by LSTM network. The CNN+LSTM model was demonstrated to perform slightly better than the single LSTM model.  \cite{Taghipour2016A}. 

\textbf{Efforts to Detect Adversarial Samples}
It is known to all that AES models are easily to be deceived by adversarial inputs if there isn't any strategy designed to help the models detect some well-designed negative samples. 
In \cite{Powers2002Stumping}, the researchers asked some experts who were familiar with e-Rater to write deceptive essays to trick e-Rater \cite{Burstein1998Automated}, which validated the fragility of existing AES systems.
As was expected, essays which were composed of some repeated well-written paragraphs achieved higher-than-deserved grades. 
To improve the robustness of AES systems, \cite{farag2018neural} utilized the window-based coherence model \cite{li2014model} to detect the adversarial crafted inputs. The results showed that their joint learning model was much more stable than other models without considering to detect adversarial samples.
The adversarial samples can roughly fall into two categories, where one is based on the well-written permuted paragraphs and has been studied by \cite{farag2018neural}, and the other represents prompt-irrelevant samples which remains to be deal with.

\section{Conclusion}
\label{conclusion}
To ensure the effectiveness and robustness of AES systems, we propose the Two-Stage Learning Framework (TSLF), which exploited both deep-encoded features and handcrafted features. In the first stage, three kinds of scores are proposed include the semantic score $S_e$, coherence score $C_e$ and prompt-relevant score $P_e$. Based on LSTM, $S_e$ is utilized to evaluate essays by taking advantage of deep semantic information, and $C_e$ and $P_e$ are applied to help AES systems to detect the adversarial samples. 
In the second stage, we concatenate the handcrafted features and these three scores, and feed the result to a boosting tree model for further training.
The results of experiments demonstrate the effectiveness of TSLF.
When compared to a number of strong baselines, our model outperforms them on five-eighths of prompts of ASAP dataset and lead the average performance. 
Besides, TSLF is much more stable when some adversarial samples are added to the test set, which is consistent with our expectation.
In summary, the effectiveness and robustness of AES system can benefit from integrating both deep-encoded features and handcrafted features.


\end{document}